\documentclass[]{fairmeta}

\usepackage{hyperref}
\usepackage{xspace}
\usepackage{listings}
\usepackage{color}
\usepackage{enumitem}
\usepackage{booktabs}
\usepackage{graphicx}
\hypersetup{breaklinks=true}
\usepackage{tablefootnote}

\definecolor{dkgreen}{rgb}{0,0.6,0}
\definecolor{gray}{rgb}{0.5,0.5,0.5}
\definecolor{mauve}{rgb}{0.58,0,0.82}

\title{TorchTitan: One-stop PyTorch native solution for production ready LLM pretraining}

\author[1]{Wanchao Liang}
\author[1]{Tianyu Liu}
\author[1]{Less Wright}
\author[1]{Will Constable}
\author[1]{Andrew Gu}
\author[1]{Chien-Chin Huang}
\author[1]{Iris Zhang}
\author[1]{Wei Feng}
\author[1]{Howard Huang}
\author[1]{Junjie Wang}
\author[2, *]{Sanket Purandare}
\author[1]{Gokul Nadathur}
\author[2]{Stratos Idreos}

\affiliation[1]{Meta}
\affiliation[2]{Harvard University}

\contribution[*]{Work done at Meta}
\newcommand{\titan}{\textsc{TorchTitan}\xspace}
\abstract{
The development of large language models (LLMs) has been instrumental in advancing state-of-the-art natural language processing applications. Training LLMs with billions of parameters and trillions of tokens require sophisticated distributed systems that enable composing and comparing several state-of-the-art techniques in order to efficiently scale across thousands of accelerators. However, existing solutions are complex, scattered across multiple libraries/repositories, lack interoperability, and are cumbersome to maintain. Thus, curating and empirically comparing training recipes require non-trivial engineering effort.

This paper introduces \titan, an open-source, PyTorch-native distributed training system that unifies and advances state-of-the-art techniques, streamlining integration and reducing engineering overhead. \titan enables seamless application of 4D parallelism in a modular and composable manner, while featuring elastic scaling to adapt to changing computational requirements. The system provides comprehensive logging, efficient checkpointing, and debugging tools, ensuring production-ready training. Moreover, \titan incorporates innovative hardware-software co-designed solutions, leveraging cutting-edge features like Float8 training and SymmetricMemory to maximize hardware utilization. As a flexible experimental test bed, \titan facilitates the curation and comparison of custom recipes for diverse training contexts. By leveraging \titan, we developed optimized training recipes for the Llama 3.1 family and provide actionable guidance on selecting and combining distributed training techniques to maximize training efficiency, based on our hands-on experiences.

We thoroughly assess \titan on the Llama 3.1 family of LLMs, spanning 8 billion to 405 billion parameters, and showcase its exceptional performance, modular composability, and elastic scalability. By stacking training optimizations, we demonstrate accelerations ranging from 65.08\% on Llama 3.1 8B at 128 GPU scale (1D), 12.59\% on Llama 3.1 70B at 256 GPU scale (2D), to 30\% on Llama 3.1 405B at 512 GPU scale (3D) on NVIDIA H100 GPUs over optimized baselines.
We also demonstrate the effectiveness of 4D parallelism in enabling long context training.
}

\date{\today}
\correspondence{Tianyu Liu at \email{lty@meta.com}}

\metadata[Code]{\url{https://github.com/pytorch/torchtitan}}

\begin{document}

\maketitle

\section{Introduction}
\label{section:intro}

Large Language Models (LLMs) \citep{devlin2018bert, liu2019optimizedbert, radford2019language, chowdhery2023palm, team2023gemini, achiam2023gpt, dubey2024llama, jiang2024mixtral, abdin2024phi} have been the driving force behind the advancement of natural language processing (NLP) applications spanning language translation, content/code generation, conversational AI, text data analysis, creative writing and art, education, and research, etc. 

Achieving state-of-the-art LLM performance requires massive scale, exemplified by top-performing models like Llama 3.1 (405B parameters, 15T tokens, 30.84M GPU hours, 16K H100 GPUs) \citep{dubey2024llama} and Google's PaLM (540B parameters, 0.8T tokens, 9.4M TPU hours, 6144 TPUv4 chips) \citep{chowdhery2023palm}. These models demonstrate exceptional natural language understanding and generation capabilities, but at the same time necessitate substantial computational resources, memory, and time to train, highlighting the significant investment required to advance natural language processing.

Training large language models (LLMs) at scale is a daunting task that requires a delicate balance of parallelism, computation, and communication, all while navigating intricate memory and computation trade-offs. The massive resources required for training make it prone to GPU failures, underscoring the need for efficient recovery mechanisms and checkpointing strategies to minimize downtime~\citep{check-nrun2022, gemini-check2023, jit-check2024, datastates2024, bytecheckpoint2024}. To optimize resource utilization and achieve elastic scalability, it is crucial to combine multiple parallelism techniques, including Data Parallel~\citep{ddp2020, zero2020, dhen-hsdp2022, fsdp2023}, Tensor Parallel~\citep{megatronlm2021, wang2022overlap, sequenceparallel2023}, Context Parallel~\citep{liu2023ring, liu2024blockwise, contextparallel2023, fang2024unifiedSP}, and Pipeline Parallel~\citep{gpipe2019, pipedream2019, megatronlm2021, zeropp2024}. By stacking these parallelisms with memory and computation optimization techniques, such as activation recomputation~\citep{chen2016trainingdeepnetssublinear, sequenceparallel2023, he2023transcending, mutwo2023}, mixed precision training~\citep{micikevicius2018mixedprecisiontraining, micikevicius2022fp8formatsdeeplearning}, and deep learning compilers~\citep{jax2018github, raf2023, onednn2024, torchcompile2024}, it is possible to maximize hardware utilization. 

While state-of-the-art distributed training techniques have significantly advanced the field, existing systems that incorporate them still fall short in addressing critical challenges that hinder their usability, adoption and effectiveness for researchers and industry practitioners.

\begin{enumerate}
    \item \emph{Non-composable}: Existing systems struggle to integrate and stack parallelism techniques, limiting multi-dimensional exploration and integration with memory and computation optimizations, thereby reducing training efficiency.
    \item \emph{Inflexible Architecture}: Lack of modularity and extensibility hampers the integration of new techniques, optimizations, and hardware, limiting adaptability to evolving ML landscapes.
    \item \emph{Inefficient Hardware Utilization}: Poor leverage of advanced hardware features results in sub-optimal GPU efficiency and lack of customizable checkpointing strategies for memory-computation trade-offs.
    \item \emph{Insufficient Support for Production Training}: Limited distributed checkpointing scalability, cumbersome failure recovery, and inadequate debugging tools hinder production-grade workflows.
    \item \emph{Framework Limitations}: Dependence on external, poorly maintained dependencies and failure to harness PyTorch’s optimized kernels, new features, and compiler support lead to inefficiencies and compatibility issues.
\end{enumerate}

The non-composability and inflexibility of distributed systems stem from the absence of unified tensor and device abstractions applied consistently across the stack. Without these foundational components, parallelism strategies, checkpointing, and efficiency optimizations remain fragmented, limiting modularity, scalability, and extensibility.

\titan’s primary research contribution lies in identifying and unifying the core principles of parallelism and optimization techniques into a cohesive framework. By leveraging and extending PyTorch’s Distributed Tensor (DTensor) and DeviceMesh~\citep{dtensor-rfc}, \titan provides a unified abstraction that simplifies the composition of parallelism strategies, and ensures correct single device semantics with its sharding primitives. Unlike existing systems that often rely on rigid or ad-hoc designs, \titan introduces a unified template for distributed training, enabling researchers to systematically explore configurations, rigorously evaluate existing methods, and uncover novel techniques within the design space.

\titan represents a complete distributed training system for large language models (LLMs), rather than merely a collection of individual techniques. Its modular, extensible architecture supports seamless composition of 4D parallelism, advanced training optimizations, and scalable distributed checkpoint save/load, all while harnessing PyTorch’s native capabilities. The system not only enable production-grade training with thousands of GPUs, but also reduces complexity and fosters innovation, setting a new standard for scalable and flexible distributed training systems.

To develop and evaluate the capabilities of \titan, we undertook several key steps, which represent the core contributions of this work, and are summarized as follows:

\begin{enumerate}
    
    \item We advance DTensor by extending its sharding to support n-D parallelism, adding compatibility with \verb|torch.compile| for compiler optimizations, and enabling  efficient checkpointing of n-D models via state dict support. We also resolve critical bugs to bolster DTensor's production readiness.
    
    \item We demonstrate how to compose various parallelism techniques, facilitating the exploration of multi-dimensional parallelism in large language model training (\S\ref{section:composability}).

    \item We enable novel hardware-software co-designed solutions exploiting advanced hardware features to increase GPU efficiency, offer customizable activation checkpointing strategies for navigating memory-computation trade-offs, and utilize \verb|torch.compile| to further optimize memory, computation, and communication (\S\ref{section:optimize}).
    
    \item We offer production grade training by incorporating scalable and efficient distributed checkpoint to facilitate fast failure recovery, integrating debugging tools like Flight Recorder to debug crashed/stuck jobs, and provide extensive logging metrics (\S\ref{section:prodready}).
    
    \item We extensively evaluate \titan on Llama 3.1 family of models, stacking 1D to 4D parallelisms (respectively), at the scale from 8 to 512 GPUs to demonstrate elastic scalability while ensuring efficiency, convergence, and accuracy. In summary, we demonstrate training accelerations ranging from 65.08\% on Llama 3.1 8B at 128 GPU scale (1D), 12.59\% on Llama3.1 70B at 256 GPU scale (2D), to 30\% on Llama3.1 405B at 512 GPU scale (3D), and the effectiveness of 4D parallelism in enabling long context training, on latest NVIDIA H100 GPUs over optimized baselines (\S\ref{section:performance}).
    
    \item We provide systematic training recipes and guidelines that empower users to navigate the complexities of distributed training, helping them optimize training efficiency for a range of model sizes and cluster configurations (\S\ref{section:guidelines}).

\end{enumerate}

By providing an accessible and extensible platform, \titan democratizes large language model (LLM) pretraining, empowering a wider range of researchers and developers to tap into the potential of LLMs and accelerate innovation in the field.

\section{Elasticity through composability}
\label{section:titan}

\begin{figure}[h!]
    \centering
    \includegraphics[width=\linewidth]{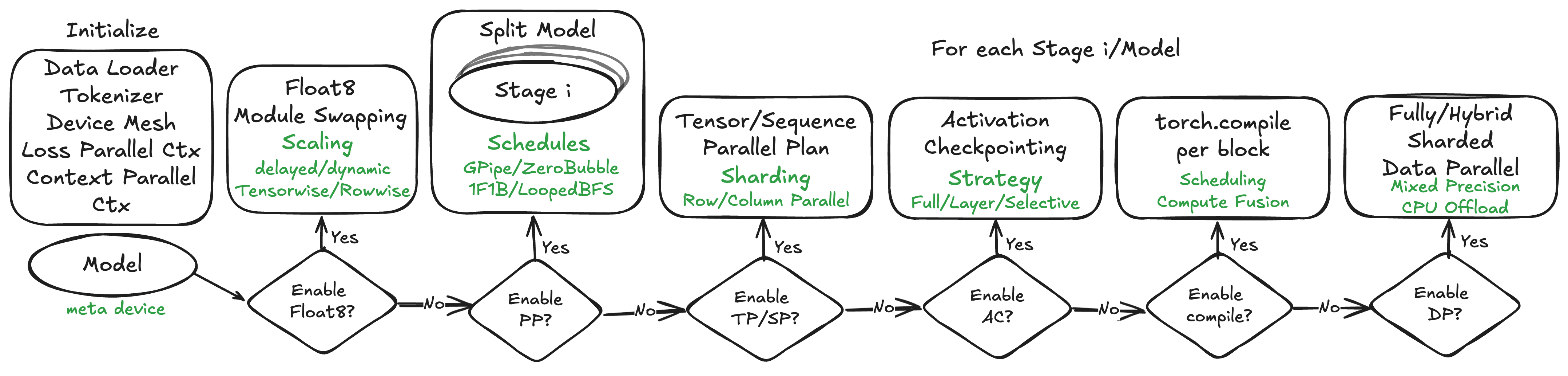}
    \caption{Composable and Modular \titan initialization workflow.}
    \label{figure:workflow}
\end{figure}

\titan incorporates various parallelisms in a modular manner to enable easy, user-selectable combinations of multi-dimensional shardings. This composability enables the tackling of difficult scaling challenges by enhancing the ease of exploration for optimizing training efficiencies at scale.

The codebase of \titan is organized purposefully to enable composability and extensibility. We intentionally keep three main components separate and as orthogonal as possible: (1) the model definition, which is parallelism-agnostic and designed for readability, (2) parallelism helpers, which apply parallelisms and training optimizations to a particular model, and (3) a generalized training loop. All these components are configurable via TOML files with command-line overrides, and it is easy to add new models and parallelism techniques on top of the existing codebase.

\subsection{Composable N-D parallelism training}
\label{section:composability}

In this section, we will walk through the entire regime of scaling model training on large clusters, including meta device initialization and the core composable multi-dimensional parallelisms, to showcase how these techniques can be composed to train LLMs efficiently at increasing scale in \titan. The corresponding code snippets in \titan can be found in Appendix~\ref{appendix:code}.

\subsubsection{Large-scale model initialization using meta device}

As LLMs grow exponentially, scaling challenges arise even before training begins, particularly in instantiating large models for sharding without exceeding CPU or GPU memory limits.

To address this, \titan enables meta device initialization, where the model is first created on a \emph{meta} device that stores only metadata, making initialization ultra-fast. The model is then sharded into Distributed Tensors (DTensors), with the local shard of each parameter residing on the meta device. Finally, parameter initialization is performed using user-defined functions, ensuring correct DTensor sharding layouts and proper RNG seed usage.

\subsubsection{Fully Sharded Data Parallel}

The original Fully Sharded Data Parallel (FSDP)~\citep{fsdp2023} is an effective implementation of ZeRO that offers large model training capability in PyTorch.   However, the original implementation (FSDP1) in PyTorch suffers from various limitations due to its FlatParameter implementation. 

Given these limitations, \titan integrates a new version of Fully Sharded Data Parallel (FSDP2), which uses the per-parameter Distributed Tensor sharding representation and thus provides better composability with model parallelism techniques and other features that require the manipulation of individual parameters.

\titan integrates and leverages FSDP2 as it's default 1D parallelism, benefiting from the improved memory management (often 7 percent lower per GPU memory requirement vs FSDP1) and the slight performance gains (average of 1.5 percent gain vs FSDP1). More details on FSDP2 and usage example are shown in Appendix~\ref{appendix:fsdp}.
 \titan makes it simple to run with FSDP2 by embedding appropriate defaults, including auto-sharding with your world size automatically. 

 For scaling to even larger world sizes, \titan also integrates Hybrid Sharded Data Parallel (HSDP) which extends FSDP2 by creating 2D DeviceMesh with replica groups.  Details are shown in Appendix \ref{appendix:hsdp}

\subsubsection{Tensor Parallel}

Tensor Parallel (TP)~\citep{megatronlm2021}, together with Sequence Parallel (SP)~\citep{sequenceparallel2023}, is a key model parallelism technique to enable large model training at scale. 

\begin{figure}[h!]
    \centering
    \includegraphics[width=0.8\linewidth]{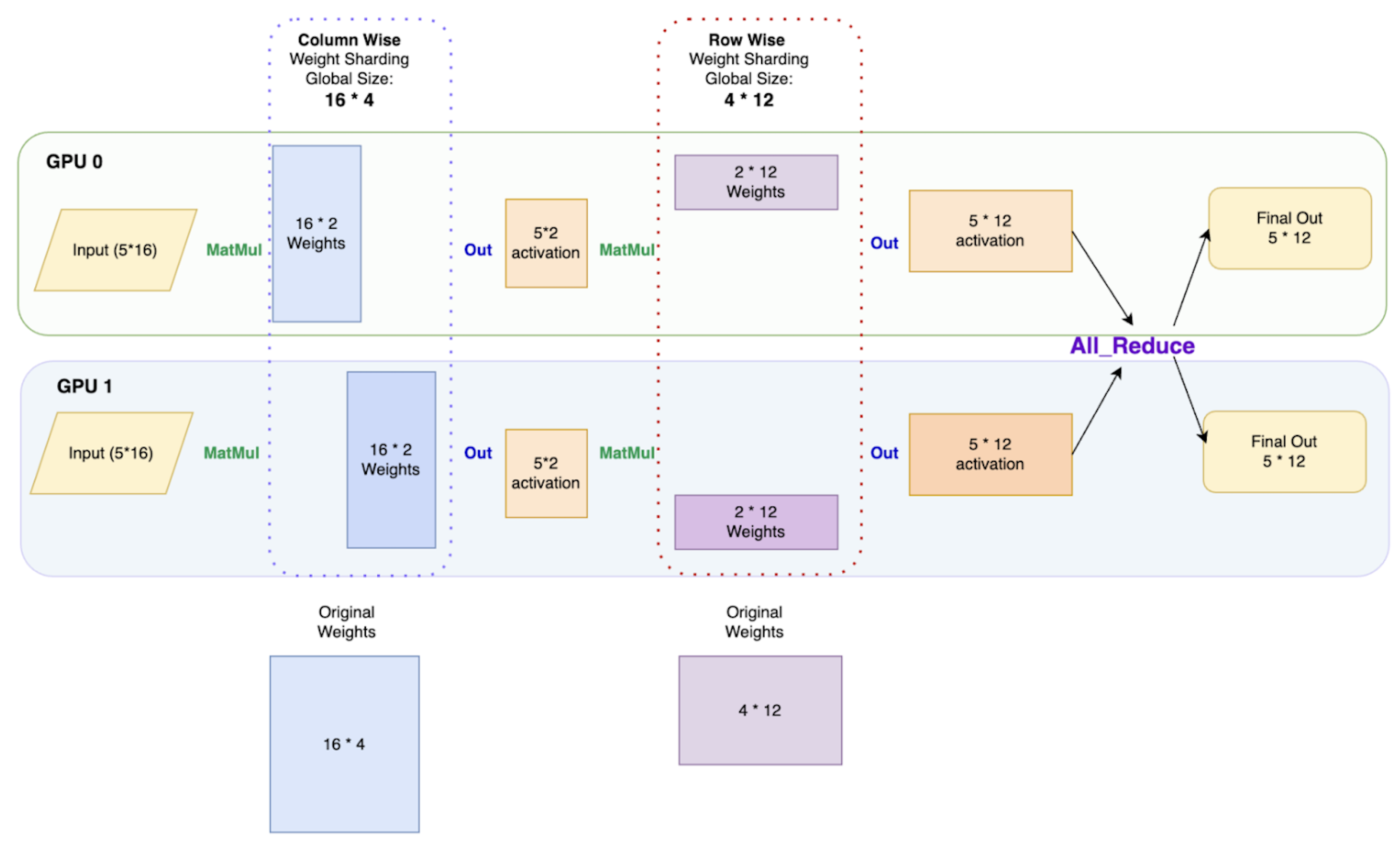}
    \caption{Tensor Parallel in detail (2 GPUs, data moves from left to right).}
    \label{figure:tp}
\end{figure}

TP is implemented in \titan using the PyTorch's \texttt{RowwiseParallel} and \texttt{ColwiseParallel} APIs, where the model parameters are partitioned to DTensors and perform sharded computation with it (Figure~\ref{figure:tp}). By leveraging DTensor, the TP implementation does not need to touch the model code, which allows faster enablement on different models and provides better composability with other features mentioned in this paper.

\paragraph{Tensor and Sequence Parallel (TP/SP)}
While TP partitions the most computationally demanding aspects, Sequence Parallel (SP) performs a sharded computation for the normalization or dropout layers on the sequence dimension, which otherwise generate large replicated activation tensors, and thus can be challenging to memory constraints per GPU. See Appendix~\ref{appendix:tp} for more details, illustrations, and usage for both TP and FSDP + TP.

Due to the synergistic relationship between TP and SP, \titan natively bundles these two together, and they are jointly controlled by the TP degree setting.

\paragraph{Loss Parallel}
When computing the loss function, model outputs are typically large, especially with TP/SP, where they are sharded across the vocabulary dimension. Naively computing cross-entropy loss requires gathering all shards, leading to high memory usage.

Loss Parallel enables efficient loss computation without fully gathering model outputs, significantly reducing memory consumption and improving training speed by minimizing communication overhead and enabling parallel sharded computation. Due to these advantages, \titan implements Loss Parallel by default. 

\subsubsection{Pipeline Parallel}

For large-scale pretraining, \titan employs Pipeline Parallelism (PP), which minimizes communication overhead by leveraging P2P communications. PP divides the model into $S$ stages, each running on a separate group of devices. Typically, each stage represents a model layer or a group of adjacent layers, but can include partial layers. During the forward pass, each stage receives input activations (except stage 0), computes locally, and sends output activations (except stage $S-1$). The last stage computes the loss and initiates the backward pass, sending gradients in reverse order. To improve efficiency, the input batch is split into microbatches, and the pipeline schedule overlaps computation and communication across microbatches. \titan supports various pipeline schedules~\citep{pipedream2019, gpipe2019, megatronlm2021, zeropp2024}.  Recently, \titan added support for new schedules including ZeroBubble and 'Flexible-Interleaved-1F1B', making use of pipeline IR to quickly express new schedules as a list of compute actions and rely on compiler passes to insert and optimize communication actions~\citealt{titan-zero-bubble}.

The PP training loop differs from standard training by creating pipeline stages and executing schedules instead of directly invoking \verb|model.forward()|. Since loss is computed per microbatch, \titan introduces a shared \verb|loss_fn| to unify pipeline and non-pipeline workflows, reducing code divergence.

\verb|torch.distributed.pipelining| also simplifies interactions with data parallelism, ensuring that reductions occur only after the final microbatch and handling shard/unshard operations (e.g., with ZeRO-3), as well as applying gradient scaling transparently within the pipeline schedule executor. For more details on \titan's implementation of PP, see Appendix~\ref{appendix:pp}.

\subsubsection{Context Parallelism}

\titan has been extended to incorporate Context Parallelism (CP)~\citep{liu2023ring, liu2024blockwise, contextparallel2023}, enabling 4D parallelism by adding CP as an additional dimension to existing DP, TP, and PP. CP scales model training by splitting the sequence dimension across GPUs, significantly increasing the maximum trainable context length without causing out-of-memory (OOM) errors. For example, on Llama 3.1 8B with 8 H100 GPUs, using CP enabled training at context lengths up to 262,144 tokens, achieving minor MFU degradation as CP degree increases~\citep{titan-cp}. For more details on CP integration please refer to Appendix~\ref{appendix:cp}.

\subsection{Optimizing training efficiencies}
\label{section:optimize}

\subsubsection{Navigating compute-memory trade-offs using activation checkpointing}

Activation checkpointing (AC)~\citep{chen2016trainingdeepnetssublinear, he2023transcending, mutwo2023} and selective activation checkpointing (SAC)~\citep{sequenceparallel2023} are standard training techniques to reduce peak GPU memory usage, by trading activation recomputation during the backward pass for memory savings. It is often needed even after applying multi-dimensional parallelisms.

\titan offers flexible AC and SAC options utilizing \verb|torch.utils.checkpoint|, applied at the \verb|TransformerBlock| level. The AC strategies include ``full'' AC,  op-level SAC, and layer-level SAC.

Within a \verb|TransformerBlock|, full AC works by recomputing all activation tensors needed during the backward pass, whereas op-level SAC saves the results from computation-intensive PyTorch operations and only recomputes others. Layer-level SAC works in similar fashion as full AC, but the wrapping is applied to every $x$ \verb|TransformerBlock| (where $x$ is specified by the user) to implement configurable trade-offs between memory and recompute.  (Details are in Appendix ~\ref{appendix:ac}.)

\subsubsection{Regional compilation to exploit \texttt{torch.compile} optimizations}

\verb|torch.compile| was released in PyTorch 2~\citep{torchcompile2024} with TorchDynamo as the frontend to extract PyTorch operations into an FX graph, and TorchInductor as the backend to compile the FX graph into fused Triton code to improve the performance.

In \titan, we use regional compilation, which applies \verb|torch.compile| to each individual \verb|TransformerBlock| in the Transformer model. This has two main benefits: 
(1) we get a full graph (without graph breaks) for each region, compatible with FSDP2 and TP (and more generally \verb|torch.Tensor| subclasses such as DTensor) and other PyTorch distributed training techniques; 
(2) since the Llama model stacks identical \verb|TransformerBlock| layers one after another, \verb|torch.compile| can identify the same structure is being repeatedly compiled and only compile once, thus greatly reducing compilation time.

\verb|torch.compile| brings efficiency in both throughput and memory (see Section~\ref{section:performance}) via computation fusions and computation-communication reordering, in a model-agnostic way with a simple user interface. Below we further elaborate how \verb|torch.compile| composability helps \titan unlock hardware-optimized performance gain with simple user interface, with the integration of advanced features such as Asynchronous TP and Float8.

\subsubsection{Asynchronous Tensor Parallel to maximally overlap communication}

By default, TP incurs blocking communications before/after the sharded computations, causing computation resources to not be effectively utilized. Asynchronous TP (AsyncTP)~\citep{wang2022overlap} achieves computation-communication overlap by fractionalizing the TP matrix multiplications within attention and feed-forward modules into smaller chunks, and overlapping communication collectives in between each section. The overlap is achieved by a micro-pipelining optimization, where results are being communicated at the same time that the other chunks of the matmul are being computed.

PyTorch AsyncTP is based on a \verb|SymmetricMemory| abstraction, which creates intra-node buffers to write faster communication collectives. This is done by allocating a shared memory buffer on each GPU in order to provide direct P2P access~\citep{titan-async-tp}.

With \titan's integration of \verb|torch.compile|, AsyncTP can be easily configured in \titan to achieve meaningful end-to-end speedups (see Section~\ref{section:performance} for details) on newer hardware (H100 or newer GPUs with NVSwitch within a node).  Usage details are in Appendix \ref{appendix:async_tp}

\subsubsection{Boosting throughput with mixed precision training and Float8 support}
Mixed precision training~\citep{micikevicius2018mixedprecisiontraining} provides both memory and computational savings while ensuring training stability.
FSDP2 has built-in support for mixed precision training with basic \verb|torch.dtype|. This covers the popular usage of performing FSDP all-gather and computation in a low precision (e.g. \verb|torch.bfloat16|), and perform lossless FSDP reduce-scatter (gradient) in high precision (e.g. \verb|torch.float32|) for better numerical results. See Appendix ~\ref{appendix:fsdp_mp} for usage details. 

\titan also supports more advanced mixed precision training with Float8, a derived data type, applied selectively to linear layers (available on newer hardware like NVIDIA H100), achieving substantial performance gains while ensuring training stability (reported in Section~\ref{section:performance}).
The Float8 feature from \verb|torchao.float8| supports multiple per-tensor scaling strategies, including dynamic, delayed, and static (see \citet{micikevicius2022fp8formatsdeeplearning, pytorch-float8}, Section 4.3 for details), while being composable with other key PyTorch-native systems such as autograd, \verb|torch.compile|, FSDP2 and TP (with Float8 all-gather capability)~\citep{titan-float8}.

\subsection{Production ready training}
\label{section:prodready}
To enable production-grade training, \titan offers seamless integration with key features out of the box. These include (1) efficient checkpointing using PyTorch Distributed Checkpointing (DCP), and (2) debugging stuck or crashed jobs through integration with Flight Recorder.

\subsubsection{Scalable and efficient Distributed Checkpointing}
Checkpoint save/load are crucial in training large language models for two reasons: they facilitate model reuse in applications like inference and evaluation, and they provide a recovery mechanism in case of failures. An optimal checkpointing workflow should ensure ease of reuse across different parallelisms and maintain high performance without slowing down training. There are two typical checkpointing methods. The first aggregates the state (model parameters and optimizer states) into an unsharded version that is parallelism-agnostic, facilitating easy reuse but requiring expensive communication. The second method has each trainer save its local sharded state, which speeds up the process but complicates reuse due to embedded parallelism information.

DCP addresses these challenges using DTensor, which encapsulates both global and local tensor information independently of parallelism. DCP converts this information into an internal format for storage. During loading, DCP matches the stored shards with the current DTensor-based model parameters and optimizer states, fetching the necessary shard from storage. \titan effectively uses DCP to balance efficiency and usability. Furthermore, DCP enhances efficiency through asynchronous checkpointing by processing storage persistence in a separate thread, allowing this operation to overlap with subsequent training iterations. \titan utilizes DCP's asynchronous checkpointing to reduce the checkpointing overhead by 5-15x compared to synchronous distributed checkpointing for the Llama 3.1 8B model~\citep{titan-dcp}.


\subsubsection{Flight Recorder to Debug Job Crashes}
Debugging NCCL collective timeouts at large scales is challenging due to the asynchronous nature of communication kernels. PyTorch’s Flight Recorder addresses this by logging the start, end, and enqueue times for all collective and p2p operations, along with metadata like process groups, source/destination ranks, tensor sizes, and stack traces.

This data is invaluable for diagnosing hangs in parallelism code. For PP, it can pinpoint the latest send or recv completed on the GPU, helping debug schedule bugs. For FSDP and TP, it identifies ranks that failed to call collectives, aiding in uncovering issues with PP scheduling or TP logic.

\section{Experimentation}
In this section, we demonstrate the effectiveness of elastic distributed training using \titan, via experiments on Llama 3.1 8B, 70B, and 405B, from 1D parallelism to 4D parallelism, at the scale from 8 GPUs to 512 GPUs.
We also share the knowledge and experience gained through \titan experimentation.
A walkthrough of the codebase on how we apply (up to) 4D parallelism can be found in Appendix~\ref{appendix:code}.

\subsection{Experimental setup}

The experiments are conducted on NVIDIA H100 GPUs\footnote{The H100 GPUs used for the experiments are non-standard. They have HBM2e and are limited to a lower TDP. The actual peak TFLOPs should be between SXM and NVL, and we don’t know the exact value.} with 95 GiB memory, where each host is equipped with 8 GPUs and NVSwitch. Two hosts form a rack connected to a TOR switch. A backend RDMA network connects the TOR switches. 
In \titan we integrate a checkpointable data loader and provide built-in support for the C4 dataset (\texttt{en} variant), a colossal, cleaned version of Common Crawl's web crawl corpus~\citep{10.5555/3455716.3455856}. We use the same dataset for all experiments in this section.
For the tokenizer, we use the official one (tiktoken) released together with Llama 3.1.

\subsection{Performance}
\label{section:performance}

To showcase the elasticity and scalability of \titan, we experiment on a wide range of GPU scales (from 8 to 512), as the underlying model size increases (8B, 70B, and 405B) with a varying number of parallelism dimensions (up to 4D).
To demonstrate the effectiveness of the optimization techniques introduced in Section~\ref{section:optimize}, we show how training throughput improves when adding each individual technique on appropriate baselines. In particular, when training on a higher dimensional parallelism with new features, the baseline is always updated to include all previous techniques.

We note that, throughout our experimentation, memory readings are stable across the whole training process\footnote{Different PP ranks can have different peak memory usages. We take the maximum across all GPUs.}, whereas throughput numbers (token per second, per GPU) are calculated and logged every 10 iterations, and always read at the (arbitrarily determined) 90th iteration.
We do not report Model FLOPS Utilization (MFU)~\citep{chowdhery2023palm} because when Float8 is enabled in \titan, both BFLOAT16 Tensor Core and FP8 Tensor Core are involved in model training, but they have different peak FLOPS and the definition of MFU under such scenario is not well-defined. We note that the 1D Llama 3.1 8B model training on 8 or 128 H100 GPUs without Float8 achieves 33\% to 42\% MFU.

\begin{table}[h!]
\begin{center}
\caption{1D parallelism (FSDP) on Llama 3.1 8B model, 8 GPUs. Mixed precision training. Selective activation checkpointing. Local batch size 2, global batch size 16. (Stats per GPU)}
\label{table:1d-8gpu}
\begin{tabular}{ l r r r r}
\toprule
 \textbf{Techniques} & \textbf{Throughput (Tok/Sec)} & \textbf{Comparison} & \textbf{Memory (GiB)} \\ 
\midrule
 FSDP & 6,258 & 100\% & 81.9 \\  
 + \verb|torch.compile| & 6,674 & + 6.64\% & 77.0 \\
 + \verb|torch.compile| + Float8 & 9,409 & + 50.35\% & 76.8 \\
\bottomrule
\end{tabular}
\end{center}
\end{table}

\begin{table}[h!]
\begin{center}
\caption{1D parallelism (FSDP) on Llama 3.1 8B model, 128 GPUs. Mixed precision training. Selective activation checkpointing. Local batch size 2, global batch size 256. (Stats per GPU)}
\label{table:1d-128gpu}
\begin{tabular}{ l r r r r}
\toprule
 \textbf{Techniques} & \textbf{Throughput (Tok/Sec)} & \textbf{Comparison} & \textbf{Memory (GiB)} \\ 
\midrule
 FSDP & 5,645 & 100\% & 67.0 \\  
 + \verb|torch.compile| & 6,482 & + 14.82\% & 62.1 \\
 + \verb|torch.compile| + Float8 & 9,319 & + 65.08\% & 61.8 \\
\bottomrule
\end{tabular}
\end{center}
\end{table}

\begin{table}[h!]
\begin{center}
\caption{2D parallelism (FSDP + TP) + \texttt{torch.compile} + Float8 on Llama 3.1 70B model, 256 GPUs. Mixed precision training. Full activation checkpointing. FSDP degree 32, TP degree 8. Local batch size 16, global batch size 512. (Stats per GPU)}
\label{table:2d-256gpu}
\begin{tabular}{ l r r r r}
\toprule
 \textbf{Techniques} & \textbf{Throughput (Tok/Sec)} & \textbf{Comparison} & \textbf{Memory (GiB)} \\ 
\midrule
 2D & 897 & 100\% & 70.3 \\  
 + AsyncTP & 1,010 & + 12.59\% & 67.7 \\
\bottomrule
\end{tabular}
\end{center}
\end{table}

\begin{table}[h!]
\begin{center}
\caption{3D parallelism (FSDP + TP + PP) + \texttt{torch.compile} + Float8 + AsyncTP on Llama 3.1 405B model, 512 GPUs. Mixed precision training. Full activation checkpointing. FSDP degree 4, TP degree 8, PP degree 16. Local batch size 32, global batch size 128. (Stats per GPU)}
\label{table:3d-512gpu}
\begin{tabular}{ l r r r r}
\toprule
 \textbf{Schedule} & \textbf{Throughput (Tok/Sec)} & \textbf{Comparison} & \textbf{Memory (GiB)} \\ 
\midrule
 1F1B & 100 & 100\% & 78.0 \\  
 Interleaved 1F1B & 130 & + 30.00\% & 80.3 \\
\bottomrule
\end{tabular}
\end{center}
\end{table}

\begin{table}[h!]
\begin{center}
\caption{FSDP + CP + \texttt{torch.compile} + Float8 on Llama 3.1 8B model, 8 GPUs. Mixed precision training. Full activation checkpointing. Local batch size 1. (Stats per GPU)}
\label{table:cp-8gpu}
\begin{tabular}{ l r r r r}
\toprule
 \textbf{Schedule} & \textbf{Sequence Length} & \textbf{Throughput (Tok/Sec)} & \textbf{Memory (GiB)} \\ 
\midrule
 FSDP 8, CP 1 & 32,768 & 3,890 & 83.9 \\
 FSDP 4, CP 2 & 65,536 & 2,540 & 84.2 \\
 FSDP 2, CP 4 & 131,072 & 1,071 & 84.0 \\
 FSDP 1, CP 8 & 262,144 & 548 & 84.5 \\
\bottomrule
\end{tabular}
\end{center}
\end{table}

\begin{table}[h!]
\begin{center}
\caption{4D parallelism (FSDP + TP + PP + CP) + \texttt{torch.compile} + Float8 + AsyncTP + 1F1B on Llama 3.1 405B model, 512 GPUs. Mixed precision training. Full activation checkpointing. TP degree 8, PP degree 8. Local batch size 8. (Stats per GPU)}
\label{table:4d-512gpu}
\begin{tabular}{ l r r r r}
\toprule
 \textbf{Schedule} & \textbf{Sequence Length} & \textbf{Throughput (Tok/Sec)} & \textbf{Memory (GiB)} \\ 
\midrule
 FSDP 8, CP 1 & 32,768 & 76 & 75.3 \\
 FSDP 4, CP 2 & 65,536 & 47 & 75.9 \\
 FSDP 2, CP 4 & 131,072 & 31 & 77.1 \\
 FSDP 1, CP 8 & 262,144 & 16 & 84.9 \\
\bottomrule
\end{tabular}
\end{center}
\end{table}

Additional experimental details and loss-convergence tests for correctness can be found in Appendix~\ref{appendix:experiment}.

\subsection {Scaling with \titan 4D Parallelism}
\label{section:guidelines}

\begin{figure}[h!]
    \centering
    \includegraphics[width=0.96\linewidth]{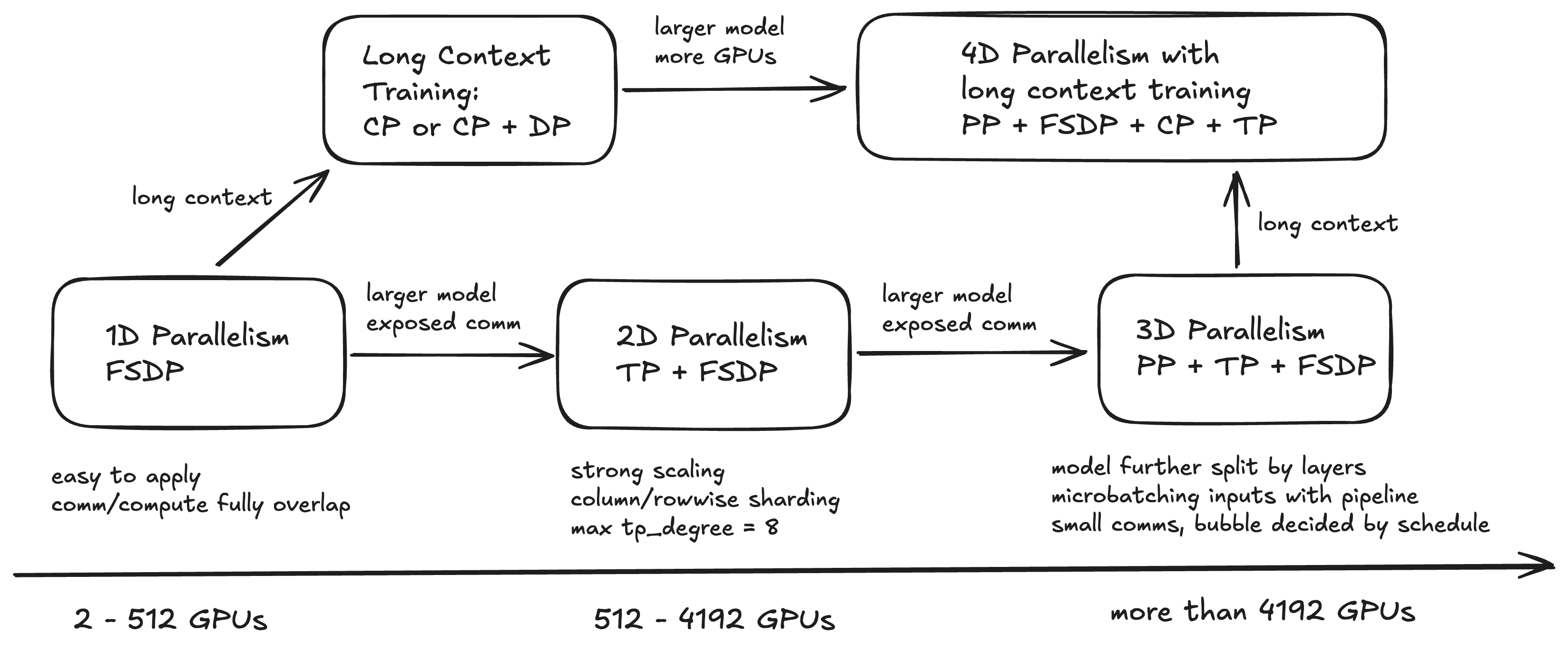}
    \caption{Scaling with 4D Parallelism}
    \label{figure:3d}
\end{figure}

Scaling large language models (LLMs) requires parallelism strategies to handle increasing model sizes and data on thousands of GPUs. \titan enables efficient scaling through composable 4D parallelism. This section highlights key observations and motivations for using \titan 4D parallelism, focusing on a specific combination shown in Figure~\ref{figure:3d}.

\subsubsection{Scaling with FSDP}
FSDP (ZeRO) is a general technique applicable to any model architecture and is often sufficient as the first degree of parallelism when communication is faster than computation (e.g., up to 512 GPUs). However, with larger scales, collective latency increases linearly with the world size, limiting efficiency. To overcome this, model parallelism like TP and PP can be combined with FSDP.

\subsubsection{2D Parallelism: TP with FSDP}
\begin{figure}[h!]
    \centering
    \includegraphics[width=0.8\linewidth]{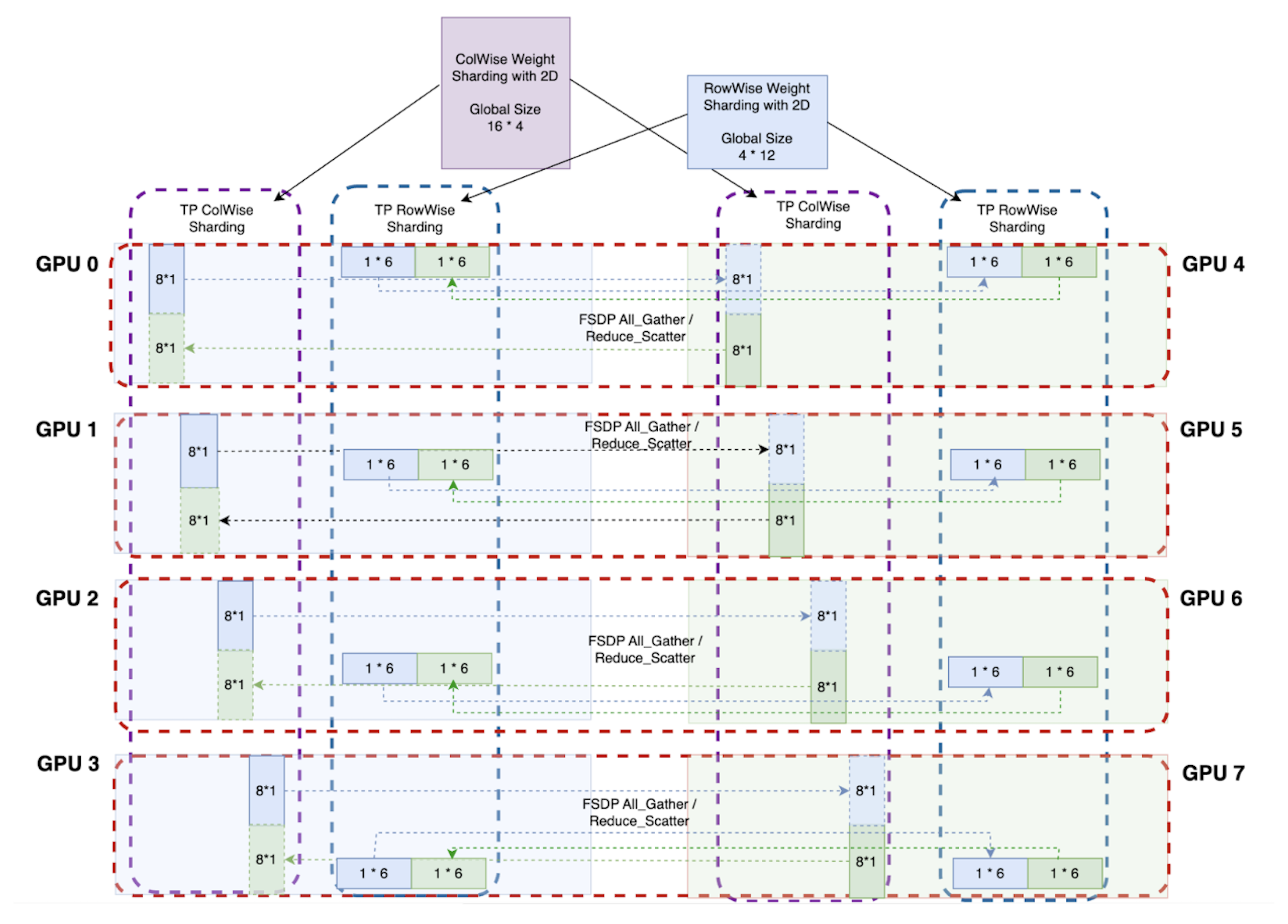}
    \caption{FSDP2 + Tensor Parallel (TP degree 4) sharding layout, with 2 nodes of 4 GPUs.}
    \label{figure:2d}
\end{figure}

Tensor Parallelism (TP) reduces collective latency by distributing work across GPUs, enabling smaller effective batch sizes and reducing peak memory usage for large models or sequence lengths. Combining FSDP and TP allows strong scaling with a fixed problem/batch size (Details shown in Figure~\ref{figure:2d}). TP also improves FLOP utilization by optimizing matrix multiplication shapes. However, TP introduces blocking collectives and is typically limited to intra-node scaling (e.g., NVLink), with degrees usually capped at 8. Scaling beyond 4192 GPUs requires combining TP with PP.

\subsubsection{3D Parallelism: PP with 2D Parallelism}
Pipeline Parallelism (PP) reduces communication bandwidth requirements by transmitting only activations and gradients between stages in a peer-to-peer manner. PP is particularly effective for mitigating FSDP communication latency at larger scales or in bandwidth-limited clusters. The efficiency of PP depends on pipeline schedules and microbatch sizes, which influence the size of pipeline ``bubbles.''

\subsubsection{Long Context Training and 4D Parallelism}
Context Parallelism (CP) allows ultra long context training by splitting the context (sequence) dimension across GPUs to avoid OOM errors. CP is mainly used for long context training, to give the model capability to capture more correlations for tokens, thus enhancing the overall model quality. For scaling sequence length, CP can be used alone or together with DP. When training large models or on large number of GPUs, we can combine CP with 3D paralleism, where TP usually keeps the innner-most DeviceMesh dimension, and CP applies in the next outer DeviceMesh dimension.

\section{Related Work}

 Libraries such as Megatron-LM~\citep{megatronlm2021}, DeepSpeed~\citep{deepspeed2020}, veScale~\citep{veScale2023} and PyTorch Distributed~\citep{pytorch2019, pytorch-distributed} provide APIs for distributed workflows. However, these frameworks present challenges in flexibility, integration, and scalability. \titan addresses these limitations with native support for key features absent in existing systems:
\begin{itemize}
    \item \emph{Megatron-LM}: Requires model modifications for TransformerEngine, lacks seamless FSDP integration with TP and PP, and does not support advanced pipeline schedules to minimize computation overhead.
    \item \emph{DeepSpeed}: Depends on Megatron-LM for TP and CP, with limited support for FSDP and advanced pipeline schedules.
    \item \emph{veScale}: Does not support FSDP, CP, SAC, Float8 training, or \texttt{torch.compile}, and offers only three pipeline schedules, compared to \titan’s six.
\end{itemize}

We note that each of these libraries has its own strengths, and \titan is designed to provide foundational components that can be leveraged by all of them. A detailed comparison, including feature breakdowns and code complexity analysis, is available in Appendix~\ref{appendix:related-work}.
Slapo~\citep{slapo2023} introduces a schedule language to convert a PyTorch model for common model training
optimizations such as 3D parallelism, and supports progressive optimization through high-level primitives. 
In contrast, \titan provides modular and composable APIs built on DTensor and DeviceMesh.


\section{Conclusion}


\titan is a powerful and flexible framework for LLM training, enabling seamless composability of parallelism techniques (FSDP, TP, PP, CP), memory optimizations (Float8, activation checkpointing), and PyTorch compiler integration for enhanced efficiency. Its modular design supports evolving architectures and hardware, fostering innovation with multi-axis metrics.

Designed for interpretability and production-grade training, \titan offers elastic scalability, comprehensive training recipes, and expert guidance on distributed training strategies. As demonstrated in experiments, it accelerates training by 65.08\% on Llama 3.1 8B (128 GPUs, 1D), 12.59\% on Llama 3.1 70B (256 GPUs, 2D), and 30\% on Llama 3.1 405B (512 GPUs, 3D) over optimized baselines, while enabling long-context training with 4D composability.
With its robust features and high efficiency, \titan is an ideal one-stop solution for challenging LLM training tasks.


\section{Acknowledgements}

We thank Soumith Chintala, Gregory Chanan, and Damien Sereni for their leadership support and product guidance. We thank Vasiliy Kuznetsov, Driss Guessous, Ke Wen, Yifu Wang, Xilun Wu, Liang Luo, and Gokul Gunasekaran for contributing fixes to the \titan repository. Finally, we would like to thank our partners Linsong Chu and Davis Wertheimer at IBM Research for evaluating \titan as a production platform and providing us with invaluable feedback.

\bibliographystyle{assets/plainnat}
\bibliography{paper}

\begin{thebibliography}{50}
\providecommand{\natexlab}[1]{#1}
\providecommand{\url}[1]{\texttt{#1}}
\expandafter\ifx\csname urlstyle\endcsname\relax
  \providecommand{\doi}[1]{doi: #1}\else
  \providecommand{\doi}{doi: \begingroup \urlstyle{rm}\Url}\fi

\bibitem[Abdin et~al.(2024)Abdin, Jacobs, Awan, Aneja, Awadallah, Awadalla, Bach, Bahree, Bakhtiari, Behl, et~al.]{abdin2024phi}
Marah Abdin, Sam~Ade Jacobs, Ammar~Ahmad Awan, Jyoti Aneja, Ahmed Awadallah, Hany Awadalla, Nguyen Bach, Amit Bahree, Arash Bakhtiari, Harkirat Behl, et~al.
\newblock Phi-3 technical report: A highly capable language model locally on your phone.
\newblock \emph{arXiv preprint arXiv:2404.14219}, 2024.

\bibitem[Achiam et~al.(2023)Achiam, Adler, Agarwal, Ahmad, Akkaya, Aleman, Almeida, Altenschmidt, Altman, Anadkat, et~al.]{achiam2023gpt}
Josh Achiam, Steven Adler, Sandhini Agarwal, Lama Ahmad, Ilge Akkaya, Florencia~Leoni Aleman, Diogo Almeida, Janko Altenschmidt, Sam Altman, Shyamal Anadkat, et~al.
\newblock {GPT}-4 technical report.
\newblock \emph{arXiv preprint arXiv:2303.08774}, 2023.

\bibitem[Anil et~al.(2023)Anil, Borgeaud, Wu, Alayrac, Yu, Soricut, Schalkwyk, Dai, Hauth, and Team]{team2023gemini}
Rohan Anil, Sebastian Borgeaud, Yonghui Wu, Jean-Baptiste Alayrac, Jiahui Yu, Radu Soricut, Johan Schalkwyk, Andrew~M Dai, Anja Hauth, and Gemini Team.
\newblock Gemini: a family of highly capable multimodal models.
\newblock \emph{arXiv preprint arXiv:2312.11805}, 2023.

\bibitem[Ansel et~al.(2024)Ansel, Yang, He, Gimelshein, Jain, Voznesensky, Bao, Bell, Berard, Burovski, Chauhan, Chourdia, Constable, Desmaison, DeVito, Ellison, Feng, Gong, Gschwind, Hirsh, Huang, Kalambarkar, Kirsch, Lazos, Lezcano, Liang, Liang, Lu, Luk, Maher, Pan, Puhrsch, Reso, Saroufim, Siraichi, Suk, Zhang, Suo, Tillet, Zhao, Wang, Zhou, Zou, Wang, Mathews, Wen, Chanan, Wu, and Chintala]{torchcompile2024}
Jason Ansel, Edward Yang, Horace He, Natalia Gimelshein, Animesh Jain, Michael Voznesensky, Bin Bao, Peter Bell, David Berard, Evgeni Burovski, Geeta Chauhan, Anjali Chourdia, Will Constable, Alban Desmaison, Zachary DeVito, Elias Ellison, Will Feng, Jiong Gong, Michael Gschwind, Brian Hirsh, Sherlock Huang, Kshiteej Kalambarkar, Laurent Kirsch, Michael Lazos, Mario Lezcano, Yanbo Liang, Jason Liang, Yinghai Lu, C.~K. Luk, Bert Maher, Yunjie Pan, Christian Puhrsch, Matthias Reso, Mark Saroufim, Marcos~Yukio Siraichi, Helen Suk, Shunting Zhang, Michael Suo, Phil Tillet, Xu~Zhao, Eikan Wang, Keren Zhou, Richard Zou, Xiaodong Wang, Ajit Mathews, William Wen, Gregory Chanan, Peng Wu, and Soumith Chintala.
\newblock {PyTorch} 2: Faster machine learning through dynamic python bytecode transformation and graph compilation.
\newblock In \emph{Proceedings of the 29th ACM International Conference on Architectural Support for Programming Languages and Operating Systems, Volume 2}, ASPLOS '24, page 929–947, New York, NY, USA, 2024. Association for Computing Machinery.
\newblock ISBN 9798400703850.
\newblock \doi{10.1145/3620665.3640366}.
\newblock \url{https://doi.org/10.1145/3620665.3640366}.

\bibitem[Bradbury et~al.(2018)Bradbury, Frostig, Hawkins, Johnson, Leary, Maclaurin, Necula, Paszke, Vander{P}las, Wanderman-{M}ilne, and Zhang]{jax2018github}
James Bradbury, Roy Frostig, Peter Hawkins, Matthew~James Johnson, Chris Leary, Dougal Maclaurin, George Necula, Adam Paszke, Jake Vander{P}las, Skye Wanderman-{M}ilne, and Qiao Zhang.
\newblock {JAX}: composable transformations of {P}ython+{N}um{P}y programs, 2018.
\newblock \url{http://github.com/jax-ml/jax}.

\bibitem[Chen et~al.(2023)Chen, Yu, Zheng, Zhang, Zhang, and Wang]{slapo2023}
Hongzheng Chen, Cody~Hao Yu, Shuai Zheng, Zhen Zhang, Zhiru Zhang, and Yida Wang.
\newblock Slapo: A schedule language for progressive optimization of large deep learning model training, 2023.
\newblock \url{https://arxiv.org/abs/2302.08005}.

\bibitem[Chen et~al.(2016)Chen, Xu, Zhang, and Guestrin]{chen2016trainingdeepnetssublinear}
Tianqi Chen, Bing Xu, Chiyuan Zhang, and Carlos Guestrin.
\newblock {Training Deep Nets with Sublinear Memory Cost}, 2016.
\newblock \url{https://arxiv.org/abs/1604.06174}.

\bibitem[Chowdhery et~al.(2023)Chowdhery, Narang, Devlin, Bosma, Mishra, Roberts, Barham, Chung, Sutton, Gehrmann, et~al.]{chowdhery2023palm}
Aakanksha Chowdhery, Sharan Narang, Jacob Devlin, Maarten Bosma, Gaurav Mishra, Adam Roberts, Paul Barham, Hyung~Won Chung, Charles Sutton, Sebastian Gehrmann, et~al.
\newblock {PaLM}: Scaling language modeling with {Pathways}.
\newblock \emph{Journal of Machine Learning Research}, 24\penalty0 (240):\penalty0 1--113, 2023.

\bibitem[Devlin(2018)]{devlin2018bert}
Jacob Devlin.
\newblock {BERT}: Pre-training of deep bidirectional {Transformers} for language understanding.
\newblock \emph{arXiv preprint arXiv:1810.04805}, 2018.

\bibitem[Dubey et~al.(2024)Dubey, Jauhri, Pandey, Kadian, Al-Dahle, Letman, Mathur, Schelten, Yang, Fan, et~al.]{dubey2024llama}
Abhimanyu Dubey, Abhinav Jauhri, Abhinav Pandey, Abhishek Kadian, Ahmad Al-Dahle, Aiesha Letman, Akhil Mathur, Alan Schelten, Amy Yang, Angela Fan, et~al.
\newblock The {Llama} 3 herd of models.
\newblock \emph{arXiv preprint arXiv:2407.21783}, 2024.

\bibitem[Eisenman et~al.(2022)Eisenman, Matam, Ingram, Mudigere, Krishnamoorthi, Nair, Smelyanskiy, and Annavaram]{check-nrun2022}
Assaf Eisenman, Kiran~Kumar Matam, Steven Ingram, Dheevatsa Mudigere, Raghuraman Krishnamoorthi, Krishnakumar Nair, Misha Smelyanskiy, and Murali Annavaram.
\newblock {Check-N-Run}: a checkpointing system for training deep learning recommendation models.
\newblock In \emph{19th USENIX Symposium on Networked Systems Design and Implementation (NSDI 22)}, pages 929--943, Renton, WA, April 2022. USENIX Association.
\newblock ISBN 978-1-939133-27-4.
\newblock \url{https://www.usenix.org/conference/nsdi22/presentation/eisenman}.

\bibitem[Fang and Zhao(2024)]{fang2024unifiedSP}
Jiarui Fang and Shangchun Zhao.
\newblock {USP}: A unified sequence parallelism approach for long context generative {AI}, 2024.
\newblock \url{https://arxiv.org/abs/2405.07719}.

\bibitem[Gupta et~al.(2024)Gupta, Krishnan, Kumar, Vijeev, Gulavani, Kwatra, Ramjee, and Sivathanu]{jit-check2024}
Tanmaey Gupta, Sanjeev Krishnan, Rituraj Kumar, Abhishek Vijeev, Bhargav Gulavani, Nipun Kwatra, Ramachandran Ramjee, and Muthian Sivathanu.
\newblock Just-in-time checkpointing: Low cost error recovery from deep learning training failures.
\newblock In \emph{Proceedings of the Nineteenth European Conference on Computer Systems}, EuroSys '24, page 1110–1125, New York, NY, USA, 2024. Association for Computing Machinery.
\newblock ISBN 9798400704376.
\newblock \doi{10.1145/3627703.3650085}.
\newblock \url{https://doi.org/10.1145/3627703.3650085}.

\bibitem[He and Yu(2023)]{he2023transcending}
Horace He and Shangdi Yu.
\newblock {Transcending runtime-memory tradeoffs in checkpointing by being fusion aware}.
\newblock \emph{Proceedings of Machine Learning and Systems}, 5:\penalty0 414--427, 2023.

\bibitem[Huang et~al.(2019)Huang, Cheng, Bapna, Firat, Chen, Chen, Lee, Ngiam, Le, Wu, and Chen]{gpipe2019}
Yanping Huang, Youlong Cheng, Ankur Bapna, Orhan Firat, Mia~Xu Chen, Dehao Chen, HyoukJoong Lee, Jiquan Ngiam, Quoc~V. Le, Yonghui Wu, and Zhifeng Chen.
\newblock \emph{GPipe: efficient training of giant neural networks using pipeline parallelism}.
\newblock Curran Associates Inc., Red Hook, NY, USA, 2019.

\bibitem[Inc.(2024)]{veScale2023}
ByteDance Inc.
\newblock {veScale}: A scalable and efficient distributed training framework.
\newblock \url{https://github.com/volcengine/veScale}, 2024.
\newblock Accessed: 2024-11-21.

\bibitem[Jiang et~al.(2024)Jiang, Sablayrolles, Roux, Mensch, Savary, Bamford, Chaplot, Casas, Hanna, Bressand, et~al.]{jiang2024mixtral}
Albert~Q Jiang, Alexandre Sablayrolles, Antoine Roux, Arthur Mensch, Blanche Savary, Chris Bamford, Devendra~Singh Chaplot, Diego de~las Casas, Emma~Bou Hanna, Florian Bressand, et~al.
\newblock Mixtral of experts.
\newblock \emph{arXiv preprint arXiv:2401.04088}, 2024.

\bibitem[Korthikanti et~al.(2023)Korthikanti, Casper, Lym, McAfee, Andersch, Shoeybi, and Catanzaro]{sequenceparallel2023}
Vijay~Anand Korthikanti, Jared Casper, Sangkug Lym, Lawrence McAfee, Michael Andersch, Mohammad Shoeybi, and Bryan Catanzaro.
\newblock Reducing activation recomputation in large transformer models.
\newblock In D.~Song, M.~Carbin, and T.~Chen, editors, \emph{Proceedings of Machine Learning and Systems}, volume~5, pages 341--353. Curan, 2023.
\newblock \url{https://proceedings.mlsys.org/paper_files/paper/2023/file/80083951326cf5b35e5100260d64ed81-Paper-mlsys2023.pdf}.

\bibitem[Li et~al.(2024)Li, Qin, Mei, Cui, Song, Chen, Zhang, Du, Cheng, Jin, Zhang, Ye, Lin, and Lavery]{onednn2024}
Jianhui Li, Zhennan Qin, Yijie Mei, Jingze Cui, Yunfei Song, Ciyong Chen, Yifei Zhang, Longsheng Du, Xianhang Cheng, Baihui Jin, Yan Zhang, Jason Ye, Eric Lin, and Dan Lavery.
\newblock {oneDNN} graph compiler: A hybrid approach for high-performance deep learning compilation.
\newblock In \emph{2024 IEEE/ACM International Symposium on Code Generation and Optimization (CGO)}, pages 460--470, 2024.
\newblock \doi{10.1109/CGO57630.2024.10444871}.

\bibitem[Li et~al.(2020)Li, Zhao, Varma, Salpekar, Noordhuis, Li, Paszke, Smith, Vaughan, Damania, et~al.]{ddp2020}
Shen Li, Yanli Zhao, Rohan Varma, Omkar Salpekar, Pieter Noordhuis, Teng Li, Adam Paszke, Jeff Smith, Brian Vaughan, Pritam Damania, et~al.
\newblock {PyTorch} distributed: Experiences on accelerating data parallel training.
\newblock \emph{arXiv preprint arXiv:2006.15704}, 2020.

\bibitem[Liu and Abbeel(2024)]{liu2024blockwise}
Hao Liu and Pieter Abbeel.
\newblock Blockwise parallel {Transformers} for large context models.
\newblock \emph{Advances in Neural Information Processing Systems}, 36, 2024.

\bibitem[Liu et~al.(2023)Liu, Zaharia, and Abbeel]{liu2023ring}
Hao Liu, Matei Zaharia, and Pieter Abbeel.
\newblock Ring attention with blockwise {Transformers} for near-infinite context.
\newblock \emph{arXiv preprint arXiv:2310.01889}, 2023.

\bibitem[Liu et~al.(2019)Liu, Ott, Goyal, Du, Joshi, Chen, Levy, Lewis, Zettlemoyer, and Stoyanov]{liu2019optimizedbert}
Yinhan Liu, Myle Ott, Naman Goyal, Jingfei Du, Mandar Joshi, Danqi Chen, Omer Levy, Mike Lewis, Luke Zettlemoyer, and Veselin Stoyanov.
\newblock {RoBERTa}: A robustly optimized {BERT} pretraining approach, 2019.
\newblock \url{https://arxiv.org/abs/1907.11692}.

\bibitem[Maurya et~al.(2024)Maurya, Underwood, Rafique, Cappello, and Nicolae]{datastates2024}
Avinash Maurya, Robert Underwood, M.~Mustafa Rafique, Franck Cappello, and Bogdan Nicolae.
\newblock Datastates-llm: Lazy asynchronous checkpointing for large language models.
\newblock In \emph{Proceedings of the 33rd International Symposium on High-Performance Parallel and Distributed Computing}, HPDC '24, page 227–239, New York, NY, USA, 2024. Association for Computing Machinery.
\newblock ISBN 9798400704130.
\newblock \doi{10.1145/3625549.3658685}.
\newblock \url{https://doi.org/10.1145/3625549.3658685}.

\bibitem[{Meta Platforms, Inc.}(2024)]{pytorch-distributed}
{Meta Platforms, Inc.}
\newblock {PyTorch Distributed}, 2024.
\newblock \url{https://pytorch.org/docs/stable/distributed.html}.
\newblock Accessed: 2023-09-26.

\bibitem[Micikevicius et~al.(2018)Micikevicius, Narang, Alben, Diamos, Elsen, Garcia, Ginsburg, Houston, Kuchaiev, Venkatesh, and Wu]{micikevicius2018mixedprecisiontraining}
Paulius Micikevicius, Sharan Narang, Jonah Alben, Gregory Diamos, Erich Elsen, David Garcia, Boris Ginsburg, Michael Houston, Oleksii Kuchaiev, Ganesh Venkatesh, and Hao Wu.
\newblock Mixed precision training, 2018.
\newblock \url{https://arxiv.org/abs/1710.03740}.

\bibitem[Micikevicius et~al.(2022)Micikevicius, Stosic, Burgess, Cornea, Dubey, Grisenthwaite, Ha, Heinecke, Judd, Kamalu, Mellempudi, Oberman, Shoeybi, Siu, and Wu]{micikevicius2022fp8formatsdeeplearning}
Paulius Micikevicius, Dusan Stosic, Neil Burgess, Marius Cornea, Pradeep Dubey, Richard Grisenthwaite, Sangwon Ha, Alexander Heinecke, Patrick Judd, John Kamalu, Naveen Mellempudi, Stuart Oberman, Mohammad Shoeybi, Michael Siu, and Hao Wu.
\newblock {FP8} formats for deep learning, 2022.
\newblock \url{https://arxiv.org/abs/2209.05433}.

\bibitem[Narayanan et~al.(2019)Narayanan, Harlap, Phanishayee, Seshadri, Devanur, Ganger, Gibbons, and Zaharia]{pipedream2019}
Deepak Narayanan, Aaron Harlap, Amar Phanishayee, Vivek Seshadri, Nikhil~R. Devanur, Gregory~R. Ganger, Phillip~B. Gibbons, and Matei Zaharia.
\newblock {PipeDream}: generalized pipeline parallelism for {DNN} training.
\newblock In \emph{Proceedings of the 27th ACM Symposium on Operating Systems Principles}, SOSP '19, page 1–15, New York, NY, USA, 2019. Association for Computing Machinery.
\newblock ISBN 9781450368735.
\newblock \doi{10.1145/3341301.3359646}.
\newblock \url{https://doi.org/10.1145/3341301.3359646}.

\bibitem[Narayanan et~al.(2021)Narayanan, Shoeybi, Casper, LeGresley, Patwary, Korthikanti, Vainbrand, Kashinkunti, Bernauer, Catanzaro, Phanishayee, and Zaharia]{megatronlm2021}
Deepak Narayanan, Mohammad Shoeybi, Jared Casper, Patrick LeGresley, Mostofa Patwary, Vijay Korthikanti, Dmitri Vainbrand, Prethvi Kashinkunti, Julie Bernauer, Bryan Catanzaro, Amar Phanishayee, and Matei Zaharia.
\newblock Efficient large-scale language model training on gpu clusters using megatron-lm.
\newblock In \emph{Proceedings of the International Conference for High Performance Computing, Networking, Storage and Analysis}, SC '21, New York, NY, USA, 2021. Association for Computing Machinery.
\newblock ISBN 9781450384421.
\newblock \doi{10.1145/3458817.3476209}.
\newblock \url{https://doi.org/10.1145/3458817.3476209}.

\bibitem[{NVIDIA}(2023)]{contextparallel2023}
{NVIDIA}.
\newblock {Megatron Core API Guide: Context Parallel}, 2023.
\newblock \url{https://docs.nvidia.com/megatron-core/developer-guide/latest/api-guide/context_parallel.html}.
\newblock Accessed: 2023-09-25.

\bibitem[Paszke et~al.(2019)Paszke, Gross, Massa, Lerer, Bradbury, Chanan, Killeen, Lin, Gimelshein, Antiga, Desmaison, K\"{o}pf, Yang, DeVito, Raison, Tejani, Chilamkurthy, Steiner, Fang, Bai, and Chintala]{pytorch2019}
Adam Paszke, Sam Gross, Francisco Massa, Adam Lerer, James Bradbury, Gregory Chanan, Trevor Killeen, Zeming Lin, Natalia Gimelshein, Luca Antiga, Alban Desmaison, Andreas K\"{o}pf, Edward Yang, Zach DeVito, Martin Raison, Alykhan Tejani, Sasank Chilamkurthy, Benoit Steiner, Lu~Fang, Junjie Bai, and Soumith Chintala.
\newblock \emph{PyTorch: an imperative style, high-performance deep learning library}.
\newblock Curran Associates Inc., Red Hook, NY, USA, 2019.

\bibitem[Purandare et~al.(2023)Purandare, Wasay, Idreos, and Jain]{mutwo2023}
Sanket Purandare, Abdul Wasay, Stratos Idreos, and Animesh Jain.
\newblock {$\mu$-TWO: 3$\texttimes$ Faster Multi-Model Training with Orchestration and Memory Optimization}.
\newblock In D.~Song, M.~Carbin, and T.~Chen, editors, \emph{Proceedings of Machine Learning and Systems}, volume~5, pages 541--562. Curan, 2023.
\newblock \url{https://proceedings.mlsys.org/paper_files/paper/2023/file/a72071d84c001596e97a2c7e1e880559-Paper-mlsys2023.pdf}.

\bibitem[{PyTorch Community}(2023)]{pytorch-float8}
{PyTorch Community}.
\newblock {Float8} in {PyTorch} 1.x, 2023.
\newblock \url{https://dev-discuss.pytorch.org/t/float8-in-pytorch-1-x/1815}.
\newblock PyTorch Discussion Thread.

\bibitem[{PyTorch Team}(2024{\natexlab{a}})]{titan-async-tp}
{PyTorch Team}.
\newblock {Introducing Async Tensor Parallelism in PyTorch}.
\newblock \url{https://discuss.pytorch.org/t/distributed-w-torchtitan-introducing-async-tensor-parallelism-in-pytorch/209487}, 2024{\natexlab{a}}.
\newblock PyTorch Forum Post.

\bibitem[{PyTorch Team}(2024{\natexlab{b}})]{titan-dcp}
{PyTorch Team}.
\newblock Optimizing checkpointing efficiency with {PyTorch DCP}.
\newblock \url{https://discuss.pytorch.org/t/distributed-w-torchtitan-optimizing-checkpointing-efficiency-with-pytorch-dcp/211250}, 2024{\natexlab{b}}.
\newblock PyTorch Forum Post.

\bibitem[{PyTorch Team}(2024{\natexlab{c}})]{titan-float8}
{PyTorch Team}.
\newblock Enabling {Float8} all-gather in {FSDP2}.
\newblock \url{https://discuss.pytorch.org/t/distributed-w-torchtitan-enabling-float8-all-gather-in-fsdp2/209323}, 2024{\natexlab{c}}.
\newblock PyTorch Forum Post.

\bibitem[{PyTorch Team}(2024{\natexlab{d}})]{titan-zero-bubble}
{PyTorch Team}.
\newblock Training with zero-bubble {Pipeline Parallelism}.
\newblock \url{https://discuss.pytorch.org/t/distributed-w-torchtitan-training-with-zero-bubble-pipeline-parallelism/214420}, 2024{\natexlab{d}}.
\newblock PyTorch Forum Post.

\bibitem[{PyTorch Team}(2025)]{titan-cp}
{PyTorch Team}.
\newblock Breaking barriers: Training long context llms with {1M} sequence length in {PyTorch} using {Context Parallel}.
\newblock \url{https://discuss.pytorch.org/t/distributed-w-torchtitan-breaking-barriers-training-long-context-llms-with-1m-sequence-length-in-pytorch-using-context-parallel/215082}, 2025.
\newblock PyTorch Forum Post.

\bibitem[Qi et~al.(2023)Qi, Wan, Huang, and Lin]{zeropp2024}
Penghui Qi, Xinyi Wan, Guangxing Huang, and Min Lin.
\newblock Zero bubble pipeline parallelism, 2023.
\newblock \url{https://arxiv.org/abs/2401.10241}.

\bibitem[Radford et~al.(2019)Radford, Wu, Child, Luan, Amodei, Sutskever, et~al.]{radford2019language}
Alec Radford, Jeffrey Wu, Rewon Child, David Luan, Dario Amodei, Ilya Sutskever, et~al.
\newblock Language models are unsupervised multitask learners.
\newblock \emph{OpenAI blog}, 1\penalty0 (8):\penalty0 9, 2019.

\bibitem[Raffel et~al.(2020)Raffel, Shazeer, Roberts, Lee, Narang, Matena, Zhou, Li, and Liu]{10.5555/3455716.3455856}
Colin Raffel, Noam Shazeer, Adam Roberts, Katherine Lee, Sharan Narang, Michael Matena, Yanqi Zhou, Wei Li, and Peter~J. Liu.
\newblock Exploring the limits of transfer learning with a unified text-to-text {Transformer}.
\newblock \emph{J. Mach. Learn. Res.}, 21\penalty0 (1), January 2020.
\newblock ISSN 1532-4435.

\bibitem[Rajbhandari et~al.(2020)Rajbhandari, Rasley, Ruwase, and He]{zero2020}
Samyam Rajbhandari, Jeff Rasley, Olatunji Ruwase, and Yuxiong He.
\newblock Zero: memory optimizations toward training trillion parameter models.
\newblock SC '20. IEEE Press, 2020.
\newblock ISBN 9781728199986.

\bibitem[Rasley et~al.(2020)Rasley, Rajbhandari, Ruwase, and He]{deepspeed2020}
Jeff Rasley, Samyam Rajbhandari, Olatunji Ruwase, and Yuxiong He.
\newblock {DeepSpeed}: System optimizations enable training deep learning models with over 100 billion parameters.
\newblock KDD '20, page 3505–3506, New York, NY, USA, 2020. Association for Computing Machinery.
\newblock ISBN 9781450379984.
\newblock \doi{10.1145/3394486.3406703}.
\newblock \url{https://doi.org/10.1145/3394486.3406703}.

\bibitem[Wan et~al.(2024)Wan, Han, Sheng, Lai, Zhang, Zhang, Peng, Lin, Liu, and Wu]{bytecheckpoint2024}
Borui Wan, Mingji Han, Yiyao Sheng, Zhichao Lai, Mofan Zhang, Junda Zhang, Yanghua Peng, Haibin Lin, Xin Liu, and Chuan Wu.
\newblock Bytecheckpoint: A unified checkpointing system for llm development, 2024.
\newblock \url{https://arxiv.org/abs/2407.20143}.

\bibitem[{Wanchao Liang}(2023)]{dtensor-rfc}
{Wanchao Liang}.
\newblock {PyTorch DTensor RFC}, 2023.
\newblock \url{https://github.com/pytorch/pytorch/issues/88838}.
\newblock GitHub Issue.

\bibitem[Wang et~al.(2022)Wang, Wei, Sabne, Davis, Ilbeyi, Hechtman, Chen, Murthy, Maggioni, Zhang, et~al.]{wang2022overlap}
Shibo Wang, Jinliang Wei, Amit Sabne, Andy Davis, Berkin Ilbeyi, Blake Hechtman, Dehao Chen, Karthik~Srinivasa Murthy, Marcello Maggioni, Qiao Zhang, et~al.
\newblock Overlap communication with dependent computation via decomposition in large deep learning models.
\newblock In \emph{Proceedings of the 28th ACM International Conference on Architectural Support for Programming Languages and Operating Systems, Volume 1}, pages 93--106, 2022.

\bibitem[Wang et~al.(2023)Wang, Jia, Zheng, Zhang, Fu, Ng, and Wang]{gemini-check2023}
Zhuang Wang, Zhen Jia, Shuai Zheng, Zhen Zhang, Xinwei Fu, T.~S.~Eugene Ng, and Yida Wang.
\newblock Gemini: Fast failure recovery in distributed training with in-memory checkpoints.
\newblock In \emph{Proceedings of the 29th Symposium on Operating Systems Principles}, SOSP '23, page 364–381, New York, NY, USA, 2023. Association for Computing Machinery.
\newblock ISBN 9798400702297.
\newblock \doi{10.1145/3600006.3613145}.
\newblock \url{https://doi.org/10.1145/3600006.3613145}.

\bibitem[Yu et~al.(2023)Yu, Fan, Huang, Jia, Liu, Wang, Zheng, Zhou, Shen, Shao, Li, and Wang]{raf2023}
Cody~Hao Yu, Haozheng Fan, Guangtai Huang, Zhen Jia, Yizhi Liu, Jie Wang, Zach Zheng, Yuan Zhou, Haichen Shen, Junru Shao, Mu~Li, and Yida Wang.
\newblock Raf: Holistic compilation for deep learning model training, 2023.
\newblock \url{https://arxiv.org/abs/2303.04759}.

\bibitem[Zhang et~al.(2022)Zhang, Luo, Liu, Li, Chen, Zhang, Wei, Hao, Tsang, Wang, Liu, Li, Badr, Park, Yang, Mudigere, and Wen]{dhen-hsdp2022}
Buyun Zhang, Liang Luo, Xi~Liu, Jay Li, Zeliang Chen, Weilin Zhang, Xiaohan Wei, Yuchen Hao, Michael Tsang, Wenjun Wang, Yang Liu, Huayu Li, Yasmine Badr, Jongsoo Park, Jiyan Yang, Dheevatsa Mudigere, and Ellie Wen.
\newblock {DHEN}: A deep and hierarchical ensemble network for large-scale click-through rate prediction, 2022.
\newblock \url{https://arxiv.org/abs/2203.11014}.

\bibitem[Zhao et~al.(2023)Zhao, Gu, Varma, Luo, Huang, Xu, Wright, Shojanazeri, Ott, Shleifer, Desmaison, Balioglu, Damania, Nguyen, Chauhan, Hao, Mathews, and Li]{fsdp2023}
Yanli Zhao, Andrew Gu, Rohan Varma, Liang Luo, Chien-Chin Huang, Min Xu, Less Wright, Hamid Shojanazeri, Myle Ott, Sam Shleifer, Alban Desmaison, Can Balioglu, Pritam Damania, Bernard Nguyen, Geeta Chauhan, Yuchen Hao, Ajit Mathews, and Shen Li.
\newblock {PyTorch FSDP}: Experiences on scaling {Fully Sharded Data Parallel}.
\newblock \emph{Proc. VLDB Endow.}, 16\penalty0 (12):\penalty0 3848–3860, aug 2023.
\newblock ISSN 2150-8097.
\newblock \doi{10.14778/3611540.3611569}.
\newblock \url{https://doi.org/10.14778/3611540.3611569}.

\end{thebibliography}

\beginappendix

\section{Composable 4D parallelism walkthrough}
\label{appendix:code}

We have discussed the scaling with \titan 4D parallelism and the motivations to apply different parallelisms to scale training to thousands of GPUs. In this section we will walk through the 4D parallelism code in \titan.

The first step is to create an instance of the model (e.g. the \verb|Transformer| for Llama models) on the meta device. We then apply PP by splitting the model into multiple PP stages according to the \verb|pipeline_parallel_split_points| config. Note that for PP with looped schedules, we may obtain multiple \verb|model_parts| from PP splitting, where each item in \verb|model_parts| is one stage-model-chunk.
Next we apply SPMD-style distributed training techniques including TP, activation checkpointing, torch.compile, FSDP, and mixed precision training for each model part, before actually initializing the sharded model on GPU.
\begin{lstlisting}
# meta init
with torch.device("meta"):
    model = model_cls.from_model_args(model_config)

# apply PP
pp_schedule, model_parts = models_pipelining_fns[model_name](
    model, pp_mesh, parallel_dims, job_config, device, model_config, loss_fn
)

for m in model_parts:
    # apply SPMD-style distributed training techniques
    models_parallelize_fns[model_name](m, world_mesh, parallel_dims, job_config)
    # move sharded model to GPU and initialize weights via DTensor
    m.to_empty(device="cuda")
    m.init_weights()
\end{lstlisting}

To apply PP to the model, we run the following code at the high level. \verb|pipeline_llama_manual_split| splits the model into multiple stages according to the manually given \verb|pipeline_parallel_split_points| config, by removing the unused model components from a complete model (on the meta device). Then \verb|build_pipeline_schedule| make the pipeline schedule with various options from \verb|torch.distributed.pipelining|, including 1F1B~\citep{pipedream2019}, GPipe~\citep{gpipe2019}, interleaved 1F1B~\citep{megatronlm2021}, etc. instructed by the \verb|pipeline_parallel_schedule| config.
\begin{lstlisting}
stages, models = pipeline_llama_manual_split(
    model, pp_mesh, parallel_dims, job_config, device, model_config
)
pp_schedule = build_pipeline_schedule(job_config, stages, loss_fn)
return pp_schedule, models
\end{lstlisting}

TP and FSDP are applied in the SPMD-style \verb|models_parallelize_fns| function.
To apply TP, we utilize the DTensor \verb|parallelize_module| API, by providing a TP ``plan'' as the instruction of how model parameters should be sharded. In the example below, we showcase the (incomplete) code for sharding the repeated \verb|TransformerBlock|.

\begin{lstlisting}
for layer_id, transformer_block in model.layers.items():
    layer_tp_plan = {
        "attention_norm": SequenceParallel(),
        "attention": PrepareModuleInput(
            input_layouts=(Shard(1), None),
            desired_input_layouts=(Replicate(), None),
        ),
        "attention.wq": ColwiseParallel(),
        ...
    }
    parallelize_module(
        module=transformer_block,
        device_mesh=tp_mesh,
        parallelize_plan=layer_tp_plan,
    )
\end{lstlisting}

Then, we apply the FSDP by wrapping each individual \texttt{TransformerBlock} and then the whole model. Note that the FSDP2 implementation in PyTorch comes with mixed precision training support. By default, we use \texttt{torch.bfloat16} on parameters all-gather and activation computations, and use \texttt{torch.float32} on gradient reduce-scatter communication and optimizer updates.

\begin{lstlisting}
mp_policy = MixedPrecisionPolicy(param_dtype, reduce_dtype)
fsdp_config = {"mesh": dp_mesh, "mp_policy": mp_policy}

for layer_id, transformer_block in model.layers.items():
    # As an optimization, do not reshard_after_forward for the last
    # TransformerBlock since FSDP would prefetch it immediately
    reshard_after_forward = int(layer_id) < len(model.layers) - 1
    fully_shard(
        transformer_block,
        **fsdp_config,
        reshard_after_forward=reshard_after_forward,
    )
fully_shard(model, **fsdp_config)
\end{lstlisting}

Independently, we can apply CP by running each training iteration under a Python context manager.
\begin{lstlisting}
optional_context_parallel_ctx = (
    utils.create_context_parallel_ctx(
        cp_mesh=world_mesh["cp"],
        cp_buffers=[input_ids, labels] + [m.freqs_cis for m in model_parts],
        cp_seq_dims=[1, 1] + [0 for _ in model_parts],
        cp_no_restore_buffers={input_ids, labels},
        cp_rotate_method=job_config.experimental.context_parallel_rotate_method,
    )
    if parallel_dims.cp_enabled
    else None
)
...
with train_context(optional_context_parallel_ctx):
    pred = model(input_ids)
    loss = loss_fn(pred, labels)
\end{lstlisting}

\section{Supplementary Materials}

\subsection{Fully Sharded Data Parallel}
\label{appendix:fsdp}
 FSDP2 advances the tensor sharding approach by replacing the original FSDP1 FlatParameter sharding. Specifically, parameters are now represented as DTensors sharded on the tensor dimension 0. This provides better composability with model parallelism techniques and other features that requires the manipulation of individual parameters, allowing sharded state dict to be represented by DTensor without any communication, and provides for a simpler meta-device initialization flow via DTensor. For example, FSDP2 unlocks finer grained tensor level quantization, especially Float8 tensor quantization, which we will showcase in the results section.

 As part of the rewrite from FSDP1 to FSDP2, FSDP2 implements an improved memory management system by avoiding using record stream.  This enables deterministic memory release, and as a result provides lower memory requirements per GPU relative to FSDP1. For example on Llama 2 7B, FSDP2 records an average of 7\% lower GPU memory versus FSDP1.  

In addition, by writing efficient kernels to perform multi-tensor allgather and reduce scatter, FSDP2 shows on-par performance compared to FSDP1, with even slight performance gains - using the Llama 2 7B, FSDP2 shows an average gain of 1.5\% faster throughput. 

The performance gains are the result of employing two small performance improvements.  First, only a single division kernel is run for the FP32 reduce scatter (pre-dividing the local FP32 reduce-scatter gradient by world size, instead of a two step pre and post divide by square root of world size). 
Secondly, in \titan, FSDP2 is integrated with a default of not re-sharding the final block in a transformer layer during the forward pass, since it will be immediately re-gathered at the start of the backward pass. 

\textbf{Usage}: \titan has fully integrated FSDP2 as the default parallelism when training, and the \verb|data_parallel_shard_degree| is the controlling dimension in the command line or TOML file. Note that for ease of use, the default \verb|data_parallel_shard_degree| is -1, means to simply use all GPUs available, so user do not need to specify the actual world size.

\subsection{Hybrid Sharded Data Parallel}
\label{appendix:hsdp}
Hybrid Sharded Data Parallel (HSDP) is an extension of FSDP \citep{dhen-hsdp2022}. In FSDP, communication occurs between all devices within the FSDP group. However, at some point, the FSDP communication overhead exceeds its corresponding computation because the latency of allgather/reduce-scatter communications increases linearly with the number of devices. This results in low MFU and becomes worthless to add more GPUs for scaling.

HSDP obviates this to some degree by creating a 2-D DeviceMesh that contains replica groups on one dimension and shard groups on the other dimension, where each shard group runs FSDP and the replica group runs normal data parallel. This ensures the FSDP communications happen in a fraction of the original world size, with the addition of backward gradient allreduce across replica groups. HSDP reduces FSDP communication overhead and allows further scaling with data parallel.

\textbf{Usage}: \titan makes it easy to experiment with HSDP by using the two configurable settings: \verb|data_parallel_shard_degree| and \verb|data_parallel_replicate_degree|, which controls the degree of the shard and replica groups we are creating. The product of both replicate and shard degree is the actual data parallel world size.

\subsection{Tensor Parallel}
\label{appendix:tp}
TP partitions the attention and feed forward network (MLP) modules of a transformer layer across multiple devices, where the number of devices used is the TP degree. This allows for multiple GPUs to cooperatively process the same batch by using the local sharded model parameters, at the cost of adding \texttt{all-reduce/all-gather/reduce-scatter} operations to synchronize intermediate activations. 

Due to the additional collectives introduced by TP, it needs to happen within a fast network (i.e NVLink). When training LLMs, TP is usually combined with FSDP, where TP shards within nodes and FSDP shards across nodes to create the 2D hierarchical sharding on different DeviceMesh dimensions. 

\textbf{Usage}: Because of the synergistic relationship between TP and SP, \titan natively bundles these two together and they are jointly controlled by the TP degree setting in the command line or the TOML entry of \verb|tensor_parallel_degree|.  Setting this to 2 for example would mean that 2 GPUs within the node will share the computational load for each transformer layers attention and MLP modules via TP, and normalization/dropout layers via Sequence Parallel. 
Loss Parallel is implemented via a context manager as it needs to control the loss computation outside of the model's forward computation. It can be enabled via \verb|enable_loss_parallel|.

\subsection{Pipeline Parallel}
\label{appendix:pp}

We expose several parameters to configure PP.  \verb|pipeline_parallel_degree| controls the number of ranks participating in PP.  \verb|pipeline_parallel_split_points| accepts a list of strings, representing layer fully-qualified-names before which a split will be performed.  Thus, the total number of pipeline stages $V$ will be determined by the length of this list. \verb|pipeline_parallel_schedule| accepts the name of the schedule to be used.  If the schedule is multi-stage, there should be $V > 1$ stages assigned to each pipeline rank, otherwise $V == 1$. 
 \verb|pipeline_parallel_microbatches| controls the number of microbatches to split a data batch into.

\subsection{Enabling 4D parallel training: Context-Parallel (CP)}
\label{appendix:cp}

To address context scaling, we have incorporated Context Parallelism (CP) into \titan.
Following the principles of modular design of \titan, CP was integrated via a context manager that dynamically replaces calls to attention operators (namely, \texttt{scaled\_dot\_product\_attention}) with CP operations, ensuring no changes to the model code are required.

Under the hood, CP shards the DTensor along the sequence dimension across the CP device mesh. It extends the DTensor dispatcher to handle CP-specific operations, such as Ring Attention and causal attention load balancing, ensuring efficient operation. By extending DTensor’s capabilities to support CP, \titan ensures that CP is fully compatible with all other parallelisms (FSDP, TP, PP), optimizations (e.g., activation checkpointing, \texttt{torch.compile}), and DCP. This demonstrates the extensibility of \titan’s modular design, which accommodates future optimizations seamlessly while maintaining performance and compatibility.

\subsection{Activation checkpointing}
 \label{appendix:ac}
 \titan offers two types of Selective Activation Checkpointing which allow for a more nuanced tradeoff between memory and recomputation. 
Specifically, we offer the option to selectively checkpoint ``per layer'' or ``per operation''.  
The goal for per operation is to free memory used by operations that are faster to recompute and save intermediates (memory) for operations that are slower to recompute and thus deliver a more effective throughput/memory trade-off.  

\textbf{Usage:}
AC is enabled via a two-line setting in the command line or TOML file.  Specifically, \verb|mode| can be either \verb|none|, \verb|selective|, or \verb|full|. 
When \verb|selective| is set, then the next config of \verb|selective_ac_type| is used which can be either a positive integer to enable selective layer checkpointing, or \verb|op| to enable selective operation checkpointing.
Per layer takes an integer input to guide the checkpointing policy, where 1 = checkpoint every layer (same as full), 2 = checkpoint every other layer, 3 = checkpoint every third layer, etc.   
Per op(eration) is driven by the \verb|_save_list| policy in \verb|parallelize_llama.py| which flags high arithmetic intensity operations such as matmul (matrix multiplication) and SPDA (Scaled Dot Product Attention) for saving the intermediate results, while allowing other lower intensity operations to be recomputed.  Note that for balancing total throughput, only every other matmul is flagged for saving.

\subsection{AsyncTP}
\label{appendix:async_tp}
The \verb|SymmetricMemory| collectives used in AsyncTP are faster than standard NCCL collectives and operate by having each GPU allocate an identical memory buffer in order to provide direct P2P access. \verb|SymmetricMemory| relies on having NVSwitch within the node, and is thus generally only available for H100 or newer GPUs.  

 \textbf{Usage}: AsyncTP is enabled within the experimental section of the \titan TOML config file and turned on or off via the \verb|enable_async_tensor_parallel| boolean setting.

 \subsection{Customizing FSDP2 Mixed Precision in \titan}
 \label{appendix:fsdp_mp}

Mixed Precision is controlled by the \verb|MixedPrecisionPolicy| class in the \verb|apply_fsdp| function, which is then customized with \verb|param_dtype| as BF16,  and \verb|reduce_dtype| defaulting to FP32 by default in \titan.    The \verb|reduce_dtype| in FP32 means that the reduce-scatter in the backwards pass for gradient computation will take place in FP32 to help maximize both stability and precision of the gradient updates.

\subsection{\titan: Comprehensive Feature Set and Reduced Complexity}
\label{appendix:related-work}

\subsubsection{\titan enables new designs}
\titan’s extensive feature set and broad design space coverage are driven by its unified design principles i.e. modularity, composability, and extensibility. Leveraging these principles, \titan seamlessly integrates diverse parallelism strategies (FSDP, TP, PP, and CP) and optimizations (e.g., SAC, Float8 training). This unified framework not only supports advanced pipeline schedules and multi-dimensional parallelism but also simplifies the integration of new techniques, making it highly adaptable for cutting-edge research and production-grade deployments.

The following table highlights \titan’s capabilities in context of parallelism, checkpointing and compiler support offerings compared to Megatron-LM, DeepSpeed, and veScale:

\begin{table}[ht]
\begin{center}
\caption{Comparison of \titan with Megatron-LM, DeepSpeed, and veScale with respect to parallelism, compiler support, activation checkpointing, and model checkpointing.}
\begin{tabular}{ l c c c c }
\toprule
\textbf{Features} & \textbf{\titan} & \textbf{Megatron-LM} & \textbf{DeepSpeed} & \textbf{veScale} \\
\midrule
FSDP-Zero2 & Yes & Yes & Yes & No \\
FSDP-Zero3 & Yes & Yes & Yes & No \\
HSDP & Yes & Yes & No & No \\
TP & Yes & Yes & No & Yes \\
Async TP (Micro-pipelining) & Yes & Yes & No & Yes \\
CP & Yes & Yes & No & No \\
PP-Gpipe & Yes & Yes & Yes & No \\
PP-Interleaved (1F1B) & Yes & Yes & Yes & Yes \\
PP-Looped-BFS & Yes & No & No & No \\
PP-1F1B & Yes & Yes & Yes & Yes \\
PP-Flexible-Interleaved-1F1B & Yes & No & No & No \\
PP-ZeroBubble & Yes & No & No & Yes \\
(TP+SP)+PP & Yes & Yes & No & Yes \\
DDP+(TP+SP)+PP & Yes & Yes & No & Yes \\
FSDP+(TP+SP) & Yes & No & No & No \\
FSDP+(TP+SP)+PP & Yes & No & No & No \\
FSDP+(TP+SP)+PP+CP & Yes & No & No & No \\
MoE & Ongoing & Yes & No & No \\
Full AC & Yes & Yes & Yes & Yes \\
Flexible SAC & Yes & No & No & No \\
DCP & Yes & Yes & Yes & Yes \\
Float8 Training & Yes & Yes & No & No \\
\verb|torch.compile| & Yes & No\tablefootnote{Custom Fusion Kernels} & Partial & No \\
\bottomrule
\end{tabular}
\label{tab:feature_comparison}
\end{center}
\end{table}

\subsubsection{Code Complexity and Maintainability}

\titan’s design principles also contribute to its significantly reduced code complexity. Despite offering a rich feature set, \titan maintains a compact and modular codebase, making it easier to extend, maintain, and evolve while ensuring high performance. The following table compares the lines of code (LOC) for \titan with Megatron-LM and DeepSpeed:

\begin{table}[h!]
\begin{center}
\caption{Lines of Code (LOC) comparison across systems.}
\begin{tabular}{ l r r r }
\toprule
\textbf{Lines of Code (LOC)} & \textbf{\titan} & \textbf{Megatron-LM} & \textbf{DeepSpeed} \\
\midrule
Core Codebase & 7K & 93K & 94K \\
Total Codebase (Including Utils) & 9K & 269K & 194K \\
\bottomrule
\end{tabular}
\label{tab:loc_comparison}
\end{center}
\end{table}

\subsection{Extended Experiments Analysis: Performance and Loss Converging}
\label{appendix:experiment}

\subsubsection{Performance}
Our experiments in Section \ref{section:performance} serve multiple objectives:
\begin{itemize}
    \item \textbf{Establish composability and modularity:} \titan demonstrates seamless integration of various parallelisms and optimization techniques.

    \item \textbf{Showcase performance improvements:} Significant speed-ups are observed across parallelisms and optimizations.

    \item \textbf{Validate elastic scalability:} \titan scales effectively with both the model size and the number of GPUs.

    \item \textbf{Ablation studies:} Detailed performance gains for individual techniques are presented.
\end{itemize}

In particular
\begin{itemize}
    \item Table \ref{table:1d-8gpu}: Highlights improvements from compiler support over eager execution, followed by further gains with Float8 training.
    \item Table \ref{table:1d-128gpu}: Demonstrates how earlier gains scale as the number of GPUs increases.
    \item Table \ref{table:2d-256gpu}: Shows speed-up achieved by AsyncTP (a HW/SW co-designed technique) over 2D training combined with \texttt{torch.compile} and Float8 training.
    \item Table \ref{table:3d-512gpu}: Quantifies the benefits of Interleaved 1F1B scheduling over 1F1B on top of AsyncTP, \texttt{torch.compile}, and Float8 training.
    \item Table \ref{table:cp-8gpu}: Demonstrates the effectiveness of CP on enabling long context training, even at small scale.
    \item Table \ref{table:4d-512gpu}: Demonstrate the composability of 4D parallelism, and the effectiveness of CP on enabling long context training at large scale.
\end{itemize}

For FSDP, the ZeRO-3 variant is used for all experiments except for those involving PP where the ZeRO-2 variant is used. This distinction is due to the inefficiency of ZeRO-3 in PP, where it incurs additional all-gather calls for each microbatch. In contrast, ZeRO-2 gathers parameters only once for the first microbatch and reshards after the last microbatch’s backward pass.

\subsubsection{Loss converging}

\titan’s design principles have influenced the development of advanced distributed training features such as FSDP2, AsyncTP, PP, and CP in PyTorch’s distributed library. Throughout these contributions, we have ensured the loss converging of individual techniques as well as their various combinations of parallelisms and optimizations.

For example, below is a series of loss-converging tests covering both parallelisms and training optimizations. We use notations of ``FSDP 8'' for an experiment in which the degree of FSDP is 8, ``FSDP 8, CP 8'' for an experiment on 64 GPUs where FSDP degree is 8 and CP degree is 8, etc.
We assume the correctness of FSDP, which can be further verified by comparing it with DDP or even single-device jobs.

\begin{table}[h!]
\begin{center}
\caption{Loss-converging tests setup.}
\label{table:loss}
\begin{tabular}{ l l l }
\toprule
 \textbf{Parallelism} & \textbf{Techniques} & \\ 
\midrule
FSDP 8 (ground truth) & default \\
FSDP 8, TP 2, PP 2 & torch.compile, Float8, async TP, Interleaved 1F1B \\
FSDP 8, TP 2, CP 2, PP 2 & torch.compile, Float8, async TP, Interleaved 1F1B \\
FSDP 8, CP 8 & default \\  
\bottomrule
\end{tabular}
\end{center}
\end{table}
\begin{figure}[t!]
    \centering
    \includegraphics[width=0.8\linewidth]{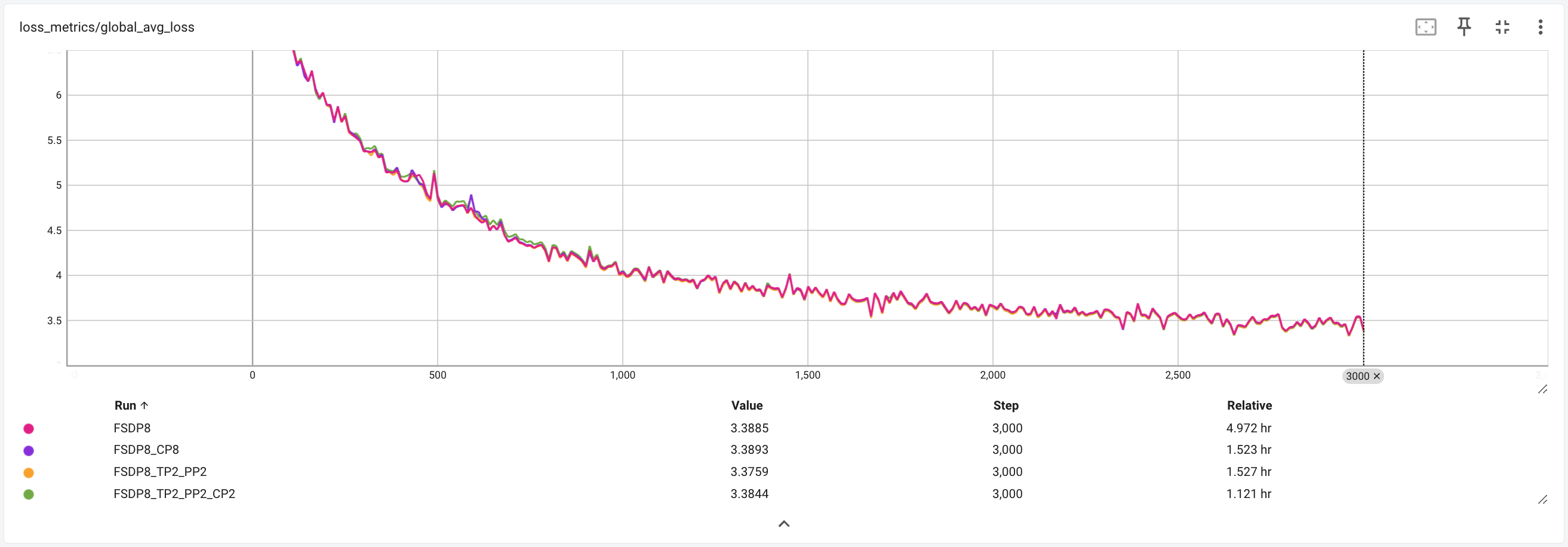}
    \caption{Loss converging tests on Llama 3.1 8B. C4 dataset. Local batch size 4, global batch size 32. 3000 steps, 600 warmup steps.}
    \label{figure:loss}
\end{figure}

\vfill

\end{document}